\documentclass[11pt,a4paper]{article}
\usepackage[hyperref]{naaclhlt2018}
\usepackage{times}
\usepackage{latexsym}
\usepackage{graphicx}
\usepackage{multicol}
\usepackage{lipsum}
\usepackage{url}
\usepackage{soulutf8}
\usepackage{subcaption}
\usepackage{enumitem}
\usepackage{xcolor}
\setlist[enumerate]{itemsep=0mm}
\usepackage{multirow}
\usepackage{tabularx}
\usepackage{booktabs}
\usepackage{bm}
\usepackage{bbm}
\usepackage{amssymb}
\usepackage{amsmath}
\usepackage{xcolor}
\newif\ifcomments
\commentstrue

\soulregister\cite7
\soulregister\ref7

\newcommand{\BCE}{\it{BCE}}
\newcommand{\MRL}{\it{MRL}}
\newcommand{\Ndata}{N^{D}}

\newcommand{\hsent}{\textsc{Hsent}}
\newcommand{\tfidf}{\textsc{TfIdf}}
\newcommand{\tfidfoh}{\textsc{TfIdf (+OH)}}

\newcommand{\wdo}{\textsc{WdO}}
\newcommand{\sent}{\textsc{Sent}}
\newcommand{\mmax}{\textsc{Max}}

\newcommand{\IM}{(A)IM}

\aclfinalcopy 

\title{Attentive Interaction Model:\\Modeling Changes in View in Argumentation}

\author{Yohan Jo$^\dagger$, Shivani Poddar$^\dagger$, Byungsoo Jeon$^\ddagger$, Qinlan Shen$^\dagger$, Carolyn P. Ros\'{e}$^\dagger$, Graham Neubig$^\dagger$\\
  $^\dagger$Language Technologies Institute, $^\ddagger$Computer Science Department \\
  School of Computer Science \\
  Carnegie Mellon University \\
  {\tt \{yohanj,spoddar2,byungsoj,qinlans,cprose,gneubig\}@cs.cmu.edu}}

\date{}

\begin{document}

\begin{verbatim}
@inproceedings{Jo:2018tt,
    author = {Jo, Yohan and Poddar, Shivani and Jeon, Byungsoo and 
              Shen, Qinlan and Ros{\'e}, Carolyn Penstein and 
              Neubig, Graham},
    title = {{Attentive Interaction Model: Modeling Changes in View 
              in Argumentation}},
    booktitle = {2018 Conference of the North American Chapter of 
                 the Association for Computational Linguistics: 
                 Human Language Technologies},
    year = {2018}
}
\end{verbatim}

\maketitle

\begin{abstract}
We present a neural architecture for modeling argumentative dialogue that explicitly models the interplay between an Opinion Holder's (OH's) reasoning and a challenger's argument, with the goal of predicting if the argument successfully changes the OH's view. The model has two components:
(1) \emph{vulnerable region detection}, an attention model that identifies parts of the OH's reasoning that are amenable to change, and (2) \emph{interaction encoding}, which identifies the relationship between the content of the OH's reasoning and that of the challenger's argument. Based on evaluation on discussions from the Change My View forum on Reddit, the two components work together to predict an OH's change in view, outperforming several baselines. A posthoc analysis suggests that sentences picked out by the attention model are addressed more frequently by successful arguments than by unsuccessful ones.\footnote{Our code is available at \url{https://github.com/yohanjo/aim}.}
\end{abstract}

\section{Introduction\label{sec:introduction}}

Through engagement in argumentative dialogue, interlocutors present arguments with the goals of winning the debate or contributing to the joint construction of knowledge. Especially modeling the knowledge co-construction process requires understanding of both the substance of viewpoints and how the substance of an argument connects with what it is arguing against. Prior work on argumentation in the NLP community, however, has focused mainly on the first goal and has often reduced the concept of a viewpoint as a discrete side (e.g., pro vs against, or liberal vs conservative), missing more nuanced and complex details of viewpoints. In addition, while the strength of the argument and the side it represents have been addressed relatively often, the dialogical aspects of argumentation have received less attention.

To bridge the gap, we present a model that jointly considers an Opinion Holder's (OH's) expressed viewpoint with a challenger's argument in order to predict if the argument succeeded in altering the OH's view.
The first component of the architecture, \textbf{vulnerable region detection}, aims to identify important parts in the OH's reasoning that are key to impacting their viewpoint. The intuition behind our model is that addressing certain parts of the OH's reasoning often has little impact in changing the OH's view, even if the OH realizes the reasoning is flawed. On the other hand, some parts of the OH's reasoning are more open to debate, and thus, it is reasonable for the model to learn and attend to parts that have a better chance to change an OH's view when addressed.  

The second component of the architecture, \textbf{interaction encoding}, aims to identify the connection between the OH's sentences and the challenger's sentences. Meaningful interaction in argumentation may include agreement/disagreement, topic relevance, or logical implication. Our model encodes the interaction between every pair of the OH's and the challenger's sentences as interaction embeddings, which are then aggregated and used for prediction. Intuitively, the interactions with the most vulnerable regions of the OH's reasoning are most critical.  Thus, in our complete model, the interaction embeddings are weighted by the vulnerability scores computed in the first component.


We evaluate our model on discussions from the Change My View forum on Reddit, where users (OHs) post their views on various issues, participate in discussion with challengers who try to change the OH's view, and acknowledge when their views have been impacted. 
Particularly, we aim to answer the following questions:
\begin{itemize}
\setlength\itemsep{0em}
    \item RQ1. Does the architecture of vulnerable region detection and interaction encoding help to predict changes in view?
    \item RQ2. Can the model identify vulnerable sentences, which are more likely to change the OH's view when addressed? If so, what properties constitute vulnerability?
    \item RQ3. What kinds of interactions between arguments are captured by the model?
\end{itemize}
We use our model to predict whether a challenger's argument has impacted the OH's view and compare the result with several baseline models.  We also present a posthoc analysis that illuminates the model's behavior in terms of vulnerable region detection and meaningful interaction.

For the remainder of the paper, we position our work in the literature (Section \ref{sec:background}) and examine the data (Section \ref{sec:data}). Then we  explain our model design (Section \ref{sec:model}). Next, we describe the experiment settings (Section \ref{sec:experiment}), discuss the results (Section \ref{sec:results}), and conclude the paper (Section \ref{sec:conclusion}).  

\section{Background\label{sec:background}}

Argumentation theories have identified important dialogical aspects of (non-)persuasive argumentation, which motivate our attempt to model the interaction of OH's and challenger's arguments. Persuasive arguments build on the hearer's accepted premises~\cite{Walton:2008tk} and appeal to emotion effectively~\cite{Aristotle:2007va}. From a challenger's perspective, effective strategies for these factors could be derived from the OH's background and reasoning. On the other hand, non-persuasive arguments may commit fallacies, such as contradicting the OH's accepted premises, diverting the discussion from the relevant and salient points suggested by the OH, failing to address the issues in question, misrepresenting the OH's reasoning, and shifting the burden of proof to the OH by asking a question~\cite{Walton:2008tk}. These fallacies can be identified only when we can effectively model how the challenger argues in relation to the OH's reasoning.

While prior work in the NLP community has studied argumentation, such as predicting debate winners~\cite{Potash:2017vu,Zhang:2016uy,Wang:2017wu,Prabhakaran:2013wv} and winning negotiation games~\cite{Keizer:2017wv}, this paper addresses a different angle: predicting whether an argument against an OH's reasoning will successfully impact the OH's view.  Some prior work investigates factors that underlie viewpoint changes~\cite{Tan:2016bk,Lukin:2017vla,Hidey:2017tt,Wei:2016ui}, but none target our task of identifying the specific arguments that impact an OH's view.

Changing an OH's view depends highly on argumentation quality, which has been the focus of much prior work.
\newcite{Wachsmuth:2017ty} reviewed theories of argumentation quality assessment and suggested a unified framework. 
Prior research has focused mainly on the presentation of an argument and some aspects in this framework without considering the OH's reasoning. Specific examples include politeness, sentiment~\cite{Tan:2016bk,Wei:2016ui}, grammaticality, factuality, topic-relatedness~\cite{Habernal:2016tg}, argument structure~\cite{Niculae:2017ura}, topics~\cite{Wang:2017wu}, and argumentative strategies (e.g., anecdote, testimony, statistics)~\cite{AlKhatib:2017ul}. Some of these aspects have been used as features to predict debate winners~\cite{Wang:2017wu} and view changes~\cite{Tan:2016bk}. \newcite{Habernal:2016um} used crowdsourcing to develop an ontology of reasons for strong/weak arguments.

The persuasiveness of an argument, however, is highly related to the OH's reasoning and how the argument connects with it. Nonetheless, research on this relationship is quite limited in the NLP community.
Existing work uses word overlap between the OH's reasoning and an argument as a feature in predicting the OH's viewpoint~\cite{Tan:2016bk}. Some studies examined the relationship between the OH's personality traits and receptivity to arguments with different topics~\cite{Ding:2016uq} or degrees of sentiment~\cite{Lukin:2017vla}. 

The most relevant to our work is the related task by \newcite{Tan:2016bk}. Their task used the same discussions from the Change My View forum as in our work and examined various stylistic features (sentiment, hedging, question marks, etc.) and word overlap features to identify discussions that impacted the OH's view. However, our task is different from theirs in that they made predictions on initial comments only, while we did so for all comments replied to by the OH in each discussion. Our task is more challenging because comments that come later in a discussion have a less direct connection to the original post. Another challenge is the extreme skew in class distribution in our data, whereas \newcite{Tan:2016bk} ensured a balance between the positive and negative classes.

The Change My View forum has received attention from recent studies. For example, \emph{ad hominem} (attacking an arguer) arguments have been studied, along with their types and causes~\cite{Habernal:2018uf}. Another study annotated semantic types of arguments and analyzed the relationship between semantic types and a change in view~\cite{Hidey:2017tt}. Although this work did not look at the interaction between OHs and specific challengers, it provides valuable insight into persuasive arguments. Additionally, the semantic types may potentially allow our model to better model complex interaction in argumentation.


\section{Data\label{sec:data}}

\begin{figure}[t]
	\centering
    \begin{subfigure}[t]{\linewidth}
        \includegraphics[width=\linewidth]{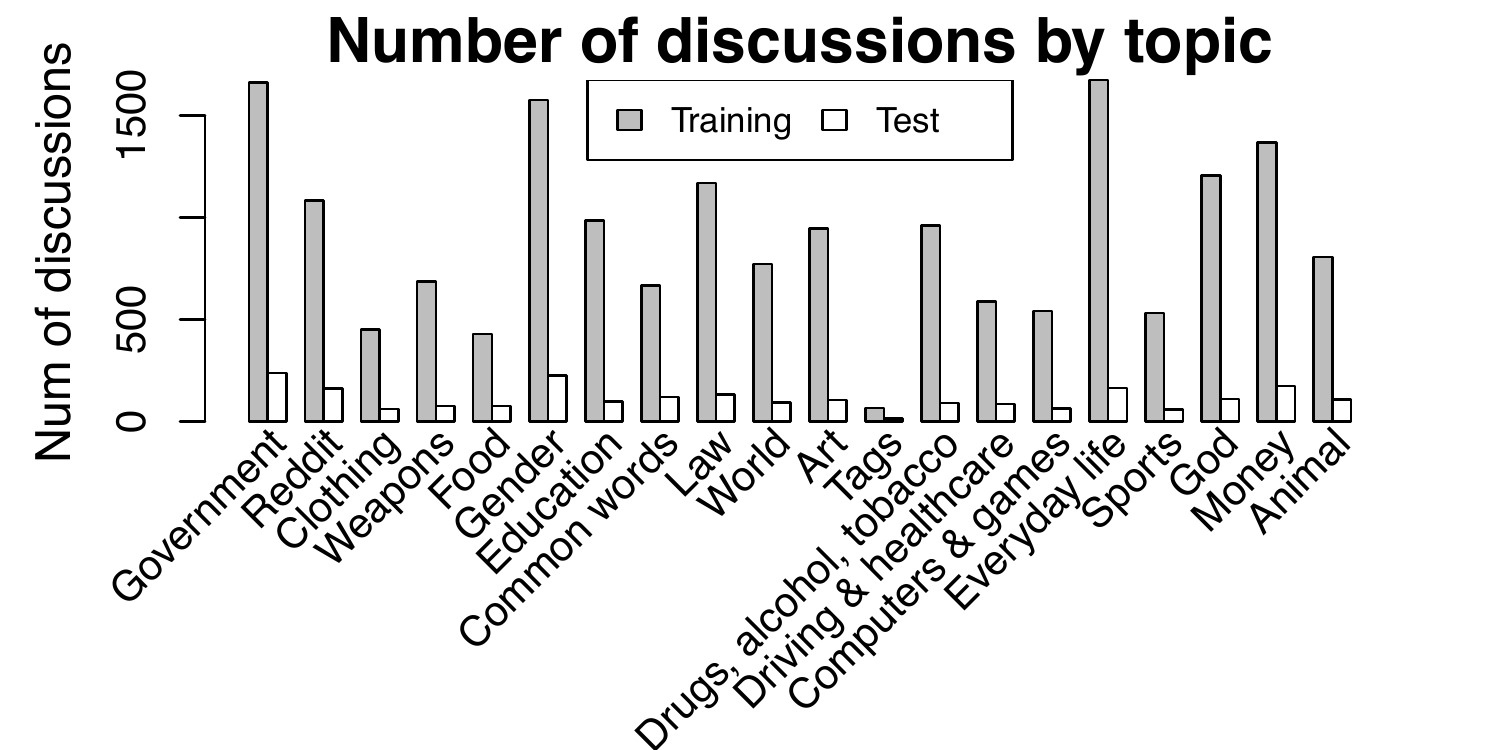}
\caption{Number of discussions by topic.\label{fig:domain_num}}
    \end{subfigure}\\
    \begin{subfigure}[t]{\linewidth}
        \includegraphics[width=\linewidth]{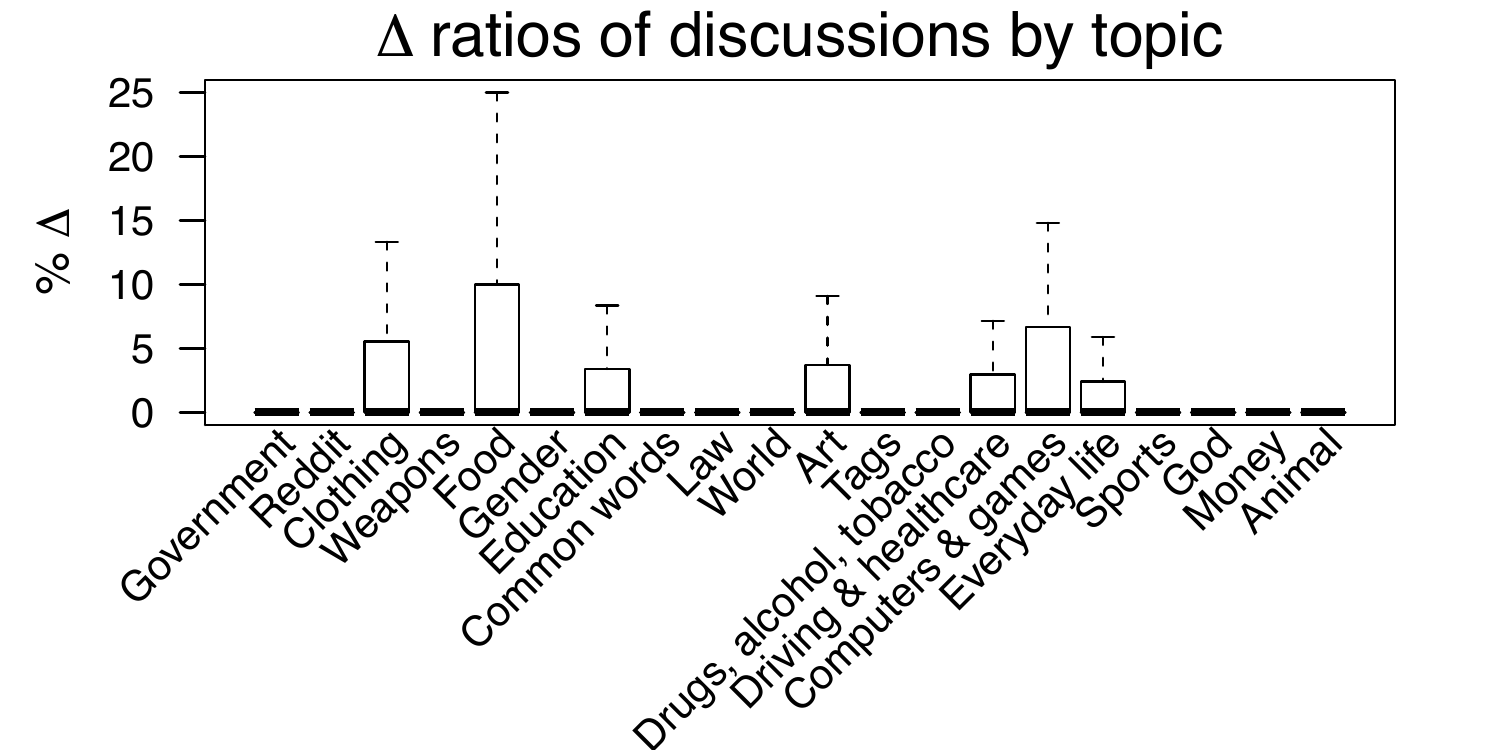}
\caption{Delta ratios in discussions by topic. (e.g., a discussion has a 10\% ratio if 10\% of the OH's replies have a $\Delta$.)\label{fig:domain_delta}}
    \end{subfigure}%
    \caption{Discussion characteristics by topic.}
\end{figure}

\begin{figure}[t]
    \centering
	\includegraphics[width=.9\linewidth]{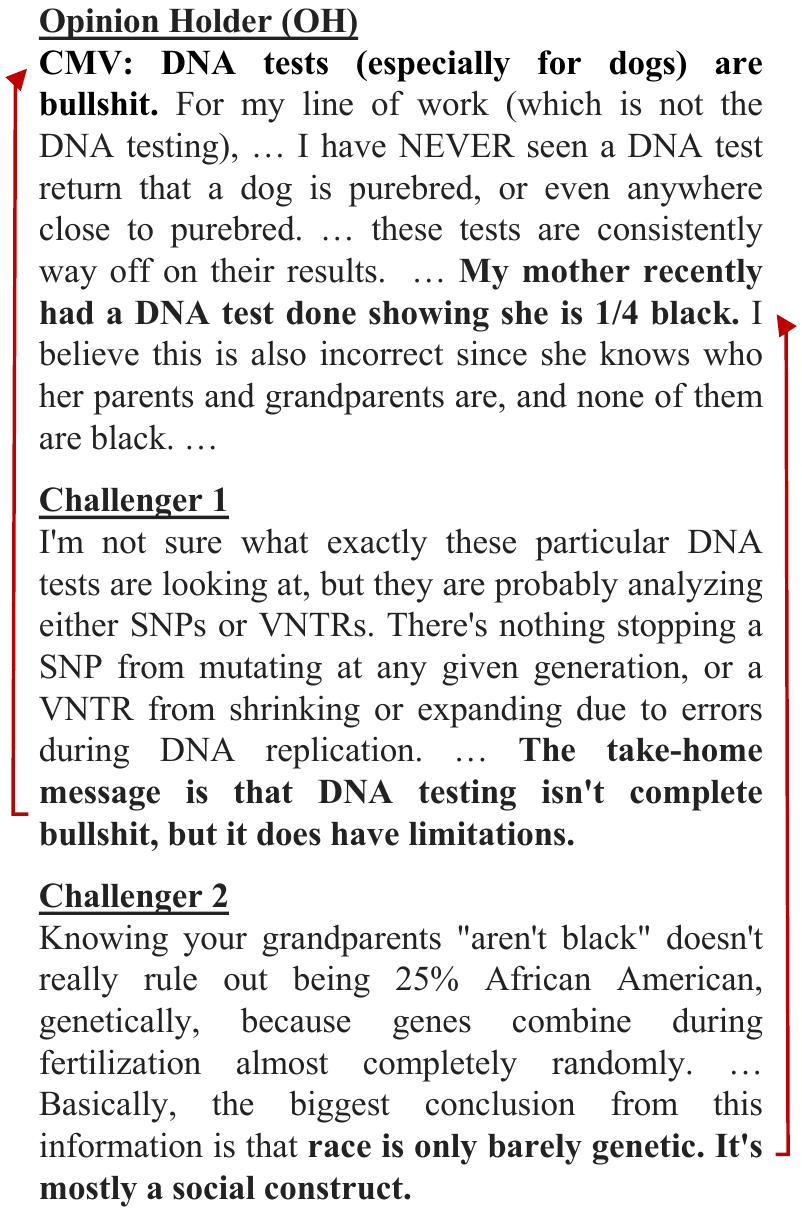}
    \caption{A discussion from Change My View.\label{fig:example_conversation}}
\end{figure}

Our study is based on discussions from the Change My View (CMV) forum\footnote{\footnotesize\url{https://www.reddit.com/r/changemyview}} on Reddit. In this forum, users (opinion holders, OHs) post their views on a wide range of issues and invite other users (challengers) to change their expressed viewpoint. If an OH gains a new insight after reading a comment, he/she replies to that comment with a $\Delta$ symbol and specifies the reasons behind his/her view change. DeltaBot monitors the forum and marks comments that received a $\Delta$, which we will use as labels indicating whether the comment successfully changed the OH's view. 

CMV discussions provide interesting insights into how people accept new information through argumentation, as OHs participate in the discussions with the explicit goal of exposing themselves to new perspectives. In addition, the rules and moderators of this forum assure high quality discussions by requiring that OHs provide enough reasoning in the initial post and replies.

We use the CMV dataset compiled by ~\newcite{Tan:2016bk}\footnote{\footnotesize\url{https://chenhaot.com/pages/changemyview.html}}. The dataset is composed of 18,363 discussions from January 1, 2013--May 7, 2015 for training data and 2,263 discussions from May 8--September 1, 2015 for test data.

\begin{figure*}[t]
\centering
\includegraphics[width=.8\linewidth]{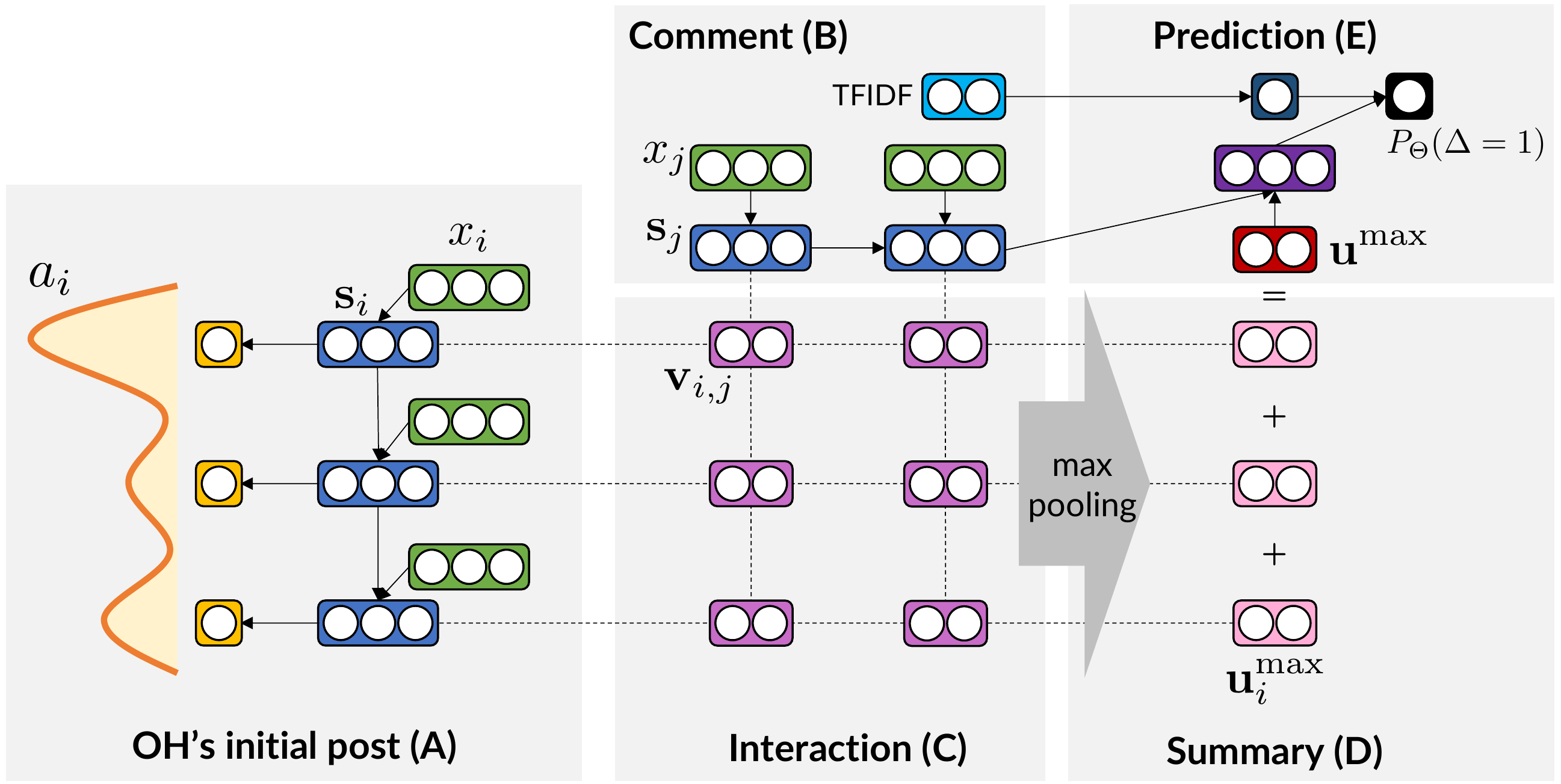}
\caption{Architecture of Attentive Interaction Model.\label{fig:model}}
\end{figure*}

\paragraph{Qualitative analysis}
We conducted qualitative analysis to better understand the data. First, to see if there are topical effects on changes in view, we examined the frequency of view changes across different topics. We ran Latent Dirichlet Allocation~\cite{Blei:2003tn} with 20 topics, taking each discussion as one document.  We assigned each discussion the topic that has the highest standardized probability. The most discussed topics are government, gender, and everyday life (Figure \ref{fig:domain_num}). As expected, the frequency of changes in view differs across topics (Figure \ref{fig:domain_delta}). The most malleable topics are food, computers \& games, clothing, art, education, and everyday life. But even in the food domain, OHs give out a $\Delta$ in less than 10\% of their replies in most discussions. 

In order to inform the design of our model, we sampled discussions not in the test set and compared comments that did and did not receive a $\Delta$. A common but often unsuccessful argumentation strategy is to correct detailed reasons and minor points of the OH's reasoning---addressing those points often has little effect, regardless of the validity of the points.  On the contrary, successful arguments usually catch incomplete parts in the OH's reasoning and offer another way of looking at an issue without threatening the OH. For instance, in the discussion in Figure \ref{fig:example_conversation}, the OH presents a negative view on DNA tests, along with his/her reasoning and experiences that justify the view. Challenger 1 addresses the OH's general statement and provides a new fact, which received a $\Delta$. On the other hand, Challenger 2 addresses the OH's issue about race but failed to change the OH's view.

When a comment addresses the OH's points,  its success relies on various interactions, including the newness of information, topical relatedness, and politeness. For example, Challenger 1 provides new information that is topically dissimilar to the OH's original reasoning. In contrast, Challenger 2's argument is relatively similar to the OH's reasoning, as it attempts to directly correct the OH's reasoning. These observations motivate the design of our Attentive Interaction Model, described in the next section.

\section{Model Specification\label{sec:model}}

Our \textbf{Attentive Interaction Model} predicts the probability of a comment changing the OH's original view, $P(\Delta = 1)$, given the OH's initial post and the comment. The architecture of the model (Figure \ref{fig:model}) consists of detecting vulnerable regions in the OH's post (sentences important to address to change the OH's view), embedding the interactions between every sentence in the OH's post and the comment, summarizing the interactions weighted by the vulnerability of OH sentences, and predicting $P(\Delta=1)$.

The main idea of our model is the architecture for capturing interactions in vulnerable regions, rather than methods for measuring specific argumentation-related features (e.g., agreement/disagreement, contraction, vulnerability, etc.). To better measure these features, we need much richer information than the dataset provides (discussion text and $\Delta$s). Therefore, our proposed architecture is not to replace prior work on argumentation features, but rather to complement it at a higher, architectural level that can potentially integrate various features. Moreover, our architecture serves as a lens for analyzing the vulnerability of OH posts and interactions with arguments. 

\newcommand{\dop}{d^{O}}
\newcommand{\xop}{x^{O}}
\newcommand{\vs}{\mathbf{s}}
\newcommand{\sop}{\vs^{O}}
\newcommand{\dco}{d^{C}}
\newcommand{\xco}{x^{C}}
\newcommand{\sco}{\vs^{C}}
\newcommand{\vv}{\mathbf{v}}
\newcommand{\vh}{\mathbf{h}}
\newcommand{\umax}{\mathbf{u}^{\rm{max}}}
\newcommand{\umean}{\mathbf{u}^{\rm{mean}}}
\newcommand{\Dse}{D^{S}}
\newcommand{\Din}{D^{I}}

\paragraph{Formal definition of the model (Figure \ref{fig:model} (A) and (B))} Denote the OH's initial post by $\dop = (\xop_1, ..., \xop_{M^{O}})$, where $x_i$ is the $i$th sentence, and $M^{O}$ is the number of sentences. The sentences are encoded via an RNN, yielding a hidden state for the $i$th sentence $\sop_i \in \mathbb{R}^{\Dse}$, where $\Dse$ is the dimensionality of the hidden states. Similarly, for a comment $\dco = (\xco_1, ..., \xco_{M^{C}})$, hidden states of the sentences $\sco_j, j = 1, \cdots, M^C$, are computed.

\paragraph{Vulnerable region detection (Figure \ref{fig:model} (A))} Given the OH's sentences, the model computes the vulnerability of the $i$th sentence $g(\sop_i) \in \mathbb{R}^1$ (e.g., using a feedforward neural network).  From this vulnerability, the attention weight of the sentence is calculated as
$$a_i = \frac{\exp g(\sop_i)}{\sum_{i' = 1}^{M^{O}} \exp g(\sop_{i'})}.$$

\paragraph{Interaction encoding (Figure \ref{fig:model} (C))} The model computes the interaction embedding of every pair of the OH's $i$th sentence and the comment's $j$th sentence,
$$\vv_{i,j} = \vh(\sop_i, \sco_j) \in \mathbb{R}^{\Din},$$
where $\Din$ is the dimensionality of interaction embeddings, and $\vh$ is an interaction function between two sentence embeddings.  $\vh$ can be a simple inner product (in which case $D^I = 1$), 
a feedforward neural network, or a more complex network. Ideally, each dimension of $\vv_{i,j}$ indicates a particular type of interaction between the pair of sentences.

\paragraph{Interaction summary (Figure \ref{fig:model} (D))} Next, for each of the OH's sentences, the model summarizes what types of meaningful interaction occur with the comment's sentences. That is, given all interaction embeddings for the OH's $i$th sentence, $\vv_{i,1}, \cdots, \vv_{i,M^C}$, the model conducts max pooling for each dimension,
\begin{align*}
\umax_i &= \left( \max_j(\vv_{i,j,1}), \cdots, \max_j(\vv_{i,j,\Din}) \right), 
\end{align*}
where $\vv_{i,j,k}$ is the $k$th dimension of $\vv_{i,j}$ and $\umax_i \in \mathbb{R}^{\Din}$. 
Intuitively, max pooling is to capture the existence of an interaction and its highest intensity for each of the OH's sentences---the interaction does not have to occur in all sentences of the comment. 
Since we have different degrees of interest in the interactions in different parts of the OH's post, we take the attention-weighted sum of $\umax_i$ to obtain the final summary vector
\begin{align*}
\umax &= \sum_{i = 1}^{M^{O}} a_i \umax_i.
\end{align*}

\paragraph{Prediction (Figure \ref{fig:model} (E))} The prediction component consists of at least one feedforward neural network, which takes as input the summary vector $\umax$ and optionally the hidden state of the last sentence in the comment $\vs_{M^C}$. 
More networks may be used to integrate other features as input, such as TFIDF-weighted $n$-grams of the comment. 
The outputs of the networks are concatenated and fed to the final prediction layer to compute $P(\Delta = 1)$. 
Using a single network that takes different kinds of features as input does not perform well, because the features are in different spaces, and linear operations between them are probably not meaningful.

\paragraph{Loss} The loss function is composed of binary cross-entropy loss and margin ranking loss. Assume there are total $\Ndata$ initial posts written by OHs, and the $l$th post has $N_l$ comments. The binary cross-entropy of the $l$th post and its $t$th comment measures the similarity between the predicted $P(\Delta = 1)$ and the true $\Delta$ as:
\begin{align*}
\BCE_{l,t} =& - \Delta_{l,t} \log P_\Theta(\Delta_{l,t} = 1) \\
&- (1 - \Delta_{l,t}) \log (1 - P_\Theta(\Delta_{l,t} = 1)),
\end{align*}
where $\Delta_{l,t}$ is the true $\Delta \in \{0, 1\}$ of the comment and $P_\Theta$ is the probability predicted by our model with parameters $\Theta$. 
Since our data is skewed to negatives, the model may overpredict $\Delta = 0$. To adjust this bias, we use margin ranking loss to drive the predicted probability of positives to be greater than the predicted probability of negatives to a certain margin. The margin ranking loss is defined on a pair of comments $C_1$ and $C_2$ with $\Delta_{C_1} > \Delta_{C_2}$ as:
\begin{align*}
&\MRL_{C_1, C_2} = \\
&~~~~~~~~ \max\{0, P_\Theta(\Delta_{C_2} = 1) - P_\Theta(\Delta_{C_1} = 1) + \epsilon \},
\end{align*}
where $\epsilon$ is a margin. Combining the two losses, our final loss is
\begin{align*}
\frac{1}{\Ndata} \sum_{l = 1}^{\Ndata} \frac{1}{N_l} \sum_{t = 1}^{N_l} \BCE_{l,t} + 
\mathbb{E}_{C_1, C_2} \left[ \MRL_{C_1, C_2} \right].
\end{align*}
For the expectation in the ranking loss, we consider all pairs of comments in each minibatch and take the mean of their ranking losses.

\section{Experiment\label{sec:experiment}}

Our task is to predict whether a comment would receive a $\Delta$, given the OH's initial post and the comment. We formulate this task as binary prediction of $\Delta \in \{0, 1\}$. Since our data is highly skewed, we use as our evaluation metric the AUC score (Area Under the Receiver Operating Characteristic Curve), which measures the probability of a positive instance receiving a higher probability of $\Delta = 1$ than a negative instance. 

\subsection{Data Preprocessing}

We exclude (1) DeltaBot's comments with no content, (2) comments replaced with \emph{[deleted]}, (3) system messages that are included in OH posts and DeltaBot's comments, (4) OH posts that are shorter than 100 characters, and (5) discussions where the OH post is excluded. We treat the title of an OH post as its first sentence. After this, every comment to which the OH replies is paired up with the OH's initial post. A comment is labeled as $\Delta = 1$ if it received a $\Delta$ and $\Delta = 0$ otherwise. Details are described in Appendix \ref{app:data}.

The original dataset comes with training and test splits (Figure \ref{fig:domain_num}). After tokenization and POS tagging with Stanford CoreNLP~\cite{manning2014stanford}, our vocabulary is restricted to the most frequent 40,000 words from the training data. For a validation split, we randomly choose 10\% of training discussions for each topic.   

\begin{table}[t]
	\centering
    \begin{tabularx}{\linewidth}{crrrr} \toprule
     & Train & Val & Test & CD\\ \midrule
    \# discussions & 4,357 & 474 & 638 & 1,548 \\
    \# pairs & 42,710 & 5,153 & 7,356 & 18,909 \\
    \# positives & 1,890 & 232 & 509 & 1,097 \\
    \bottomrule\end{tabularx}
    \caption{Data statistics. (CD: cross-domain test)\label{tab:data_statistics}}
\end{table}

We train our model on the seven topics that have the highest $\Delta$ ratios (Figure \ref{fig:domain_delta}). We test on the same set of topics for in-domain evaluation and on the other 13 topics for cross-domain evaluation. The main reason for choosing the most malleable topics is that these topics provide more information about people learning new perspectives, which is the focus of our paper. Some statistics of the resulting data are in Table \ref{tab:data_statistics}.

\subsection{Inputs}
We use two basic types of inputs: sentence embeddings and TFIDF vectors. These basic inputs are by no means enough for our complex task, and most prior work utilizes higher-level features (politeness, sentiment, etc.) and task-specific information. Nevertheless, our experiment is limited to the basic inputs to minimize feature engineering and increase replicability, but our model is general enough to incorporate other features as well.

\paragraph{Sentence embeddings} Our input sentences $x$ are sentence embeddings obtained by a pretrained sentence encoder~\cite{Conneau:2017uf} (this is different from the sentence encoder in our model). The pretrained sentence encoder is a BiLSTM with max pooling trained on the Stanford Natural Language Inference corpus~\cite{snli:emnlp2015} for textual entailment. Sentence embeddings from this encoder, combined with logistic regression on top, showed good performance in various transfer tasks, such as entailment and caption-image retrieval ~\cite{Conneau:2017uf}. 

\paragraph{TFIDF}
A whole post or comment is represented as a TFIDF-weighted bag-of-words, where IDF is based on the training data. We consider the top 40,000 $n$-grams ($n = 1, 2, 3$) by term frequency.

\paragraph{Word Overlap}
Although integration of hand-crafted features is behind the scope of this paper, we test the word overlap features between a comment and the OH's post, introduced by \newcite{Tan:2016bk}, as simple proxy for the interaction. For each comment, given the set of its words $C$ and that of the OH's post $O$, these features are defined as $\left[ |C \cap O|, \frac{|C \cap O|}{|C|}, \frac{|C \cap O|}{|O|}, \frac{|C \cap O|}{|C \cup O|} \right]$.

\subsection{Model Setting}

\paragraph{Network configurations} For sentence encoding, Gated Recurrent Units~\cite{Cho:2014ua} with hidden state sizes 128 or 192 are explored.
For attention, a single-layer feedforward neural network (FF) with one output node is used. For interaction encoding, we explore two interaction functions: (1) the inner product of the sentence embeddings and (2) a two-layer FF with 60 hidden nodes and three output nodes with a concatenation of the sentence embeddings as input. For prediction, we explore (1) a single-layer FF with either one output node if the summary vector $\umax$ is the only input or 32 or 64 output nodes with ReLU activation if the hidden state of the comment's last sentence is used as input, and optionally (2) a single-layer FF with 1 or 3 output nodes with ReLU activation for the TFIDF-weighted $n$-grams of the comment.
The final prediction layer is a single-layer FF with one output node with sigmoid activation that takes the outputs of the two networks above and optionally the word overlap vector. 
The margin $\epsilon$ for the ranking margin loss is 0.5. Optimization is performed using AdaMax with the initial learning rate 0.002, decayed by 5\% every epoch. Training stops after 10 epochs if the average validation AUC score of the last 5 epochs is lower than that of the first 5 epochs; otherwise, training runs 5 more epochs. The minibatch size is 10.

\paragraph{Input configurations} The prediction component of the model takes combinations of the inputs: \mmax{} ($\umax$), 
\hsent{} (the last hidden state of the sentence encoder $\sco_{M^C}$), \tfidf{} (TFIDF-weighted $n$-grams of the comment), and \wdo{} (word overlap).

\subsection{Baseline\label{sec:baselines}}
The most similar prior work to ours~ \cite{Tan:2016bk} predicted whether an OH would ever give a $\Delta$ in a discussion. The work used logistic regression with bag-of-words features. Hence, we also use logistic regression as our baseline to predict $P(\Delta = 1)$. Simple logistic regression using TFIDF is a relatively strong baseline, as it beat more complex features in the aforementioned task.

\paragraph{Model configurations} Different regularization methods (L1, L2),  regularization strengths ($2^{\wedge}\{-1, 0, 1, 2\}$), and class weights for positives (1, 2, 5) are explored. Class weights penalize false-negatives differently from false-positives, which is appropriate for the skewed data. 

\paragraph{Input configurations} The model takes combinations of the inputs: \tfidf{} (TFIDF-weighted $n$-grams of the comment), \tfidfoh{} (concatenation of the TFIDF-weighted $n$-grams of the comment and the OH's post), \wdo{} (word overlap), and \sent{} (the sum of the input sentence embeddings of the comment).

\section{Results\label{sec:results}}

\begin{table}[t]
    \centering
    \begin{tabularx}{\linewidth}{cccc} \toprule
    Model & Inputs & ID & CD \\ \midrule
    LR & \sent{} & 62.8 & 62.5  \\
    LR & \tfidfoh{} & 69.5 & 69.1  \\
    LR & \tfidf{} & 70.9 & \textbf{69.6}  \\
    LR & \sent{}+\tfidf{} & 64.0 & 63.1 \\ 
    LR & \tfidf{}+\wdo{} & 71.1 & 69.5 \\ \midrule
    AIM & \mmax{} & 70.5 & 67.5 \\
    AIM & \mmax{}+\tfidf{} & \textbf{\ 72.0*} & 69.4 \\ 
    AIM & \mmax{}+\tfidf{}+\wdo{} & 70.9 & 68.4 \\\midrule
    \IM & \hsent{} & 69.6 & 67.6 \\
    \IM & \hsent{}+\tfidf{} & 69.0 & 67.6 \\
    \IM & \mmax{}+\tfidf{} & 69.5 & 68.1 \\ \bottomrule
    \end{tabularx}
    \caption{AUC scores. (ID: in-domain AUC (\%), CD: cross-domain AUC (\%), LR: logistic regression, AIM: Attention Interaction Model, \IM: AIM without attention.) *: $p < 0.05$ using the DeLong test compared to LR with \tfidf{}. \label{tab:baseline_accuracy}}
\end{table}

Table \ref{tab:baseline_accuracy} shows the test AUC scores for the baseline and our model in different input configurations. For each configuration, we chose the optimal parameters based on validation AUC scores.

\paragraph{RQ1. Does the architecture of vulnerable region detection and interaction encoding help to predict changes in view?}
Both interaction information learned by our model and surface-level $n$-grams in \tfidf{} have strong predictive power, and attending to vulnerable regions helps. The highest score is achieved by our model (AIM) with both \mmax{} and \tfidf{} as input (72.0\%). The performance drops if the model does not use interaction information---\IM{} with \hsent{} (69.6\%)---or vulnerability information---\IM{} with \mmax{}+\tfidf{} (69.5\%). 

\tfidf{} by itself is also a strong predictor, as logistic regression with \tfidf{} performs well (70.9\%).
There is a performance drop if \tfidf{} is not used in most settings. This is unsurprising because TFIDF captures some topical or stylistic information that was shown to play important roles in argumentation in prior work~\cite{Tan:2016bk,Wei:2016ui}. Simply concatenating both comment's and OH's TFIDF features does not help (69.5\%), most likely due to the fact that a simple logistic regression does not capture interactions between features.

When the hand-crafted word overlap features are integrated to LR, the accuracy is increased slightly, but the difference is not statistically significant compared to LR without these features nor to the best AIM configuration. These features do not help AIM (70.9\%), possibly because the information is redundant, or AIM requires a more deliberate way of integrating hand-crafted features.

For cross-domain performance, logistic regression with \tfidf{} performs best (69.6\%). Our interaction information does not transfer to unseen topics as well as \tfidf{}. This weakness is alleviated when our model uses \tfidf{} in addition to \mmax{}, increasing the cross-domain score (from 67.5\% to 69.4\%).
We expect that information about vulnerability would have more impact within domain than across domains because it may learn domain-specific information about which kinds of reasoning are vulnerable.

The rest of the section reports our qualitative analysis based on the best model configuration.

\begin{figure*}
    \centering
    \includegraphics[width=\linewidth]{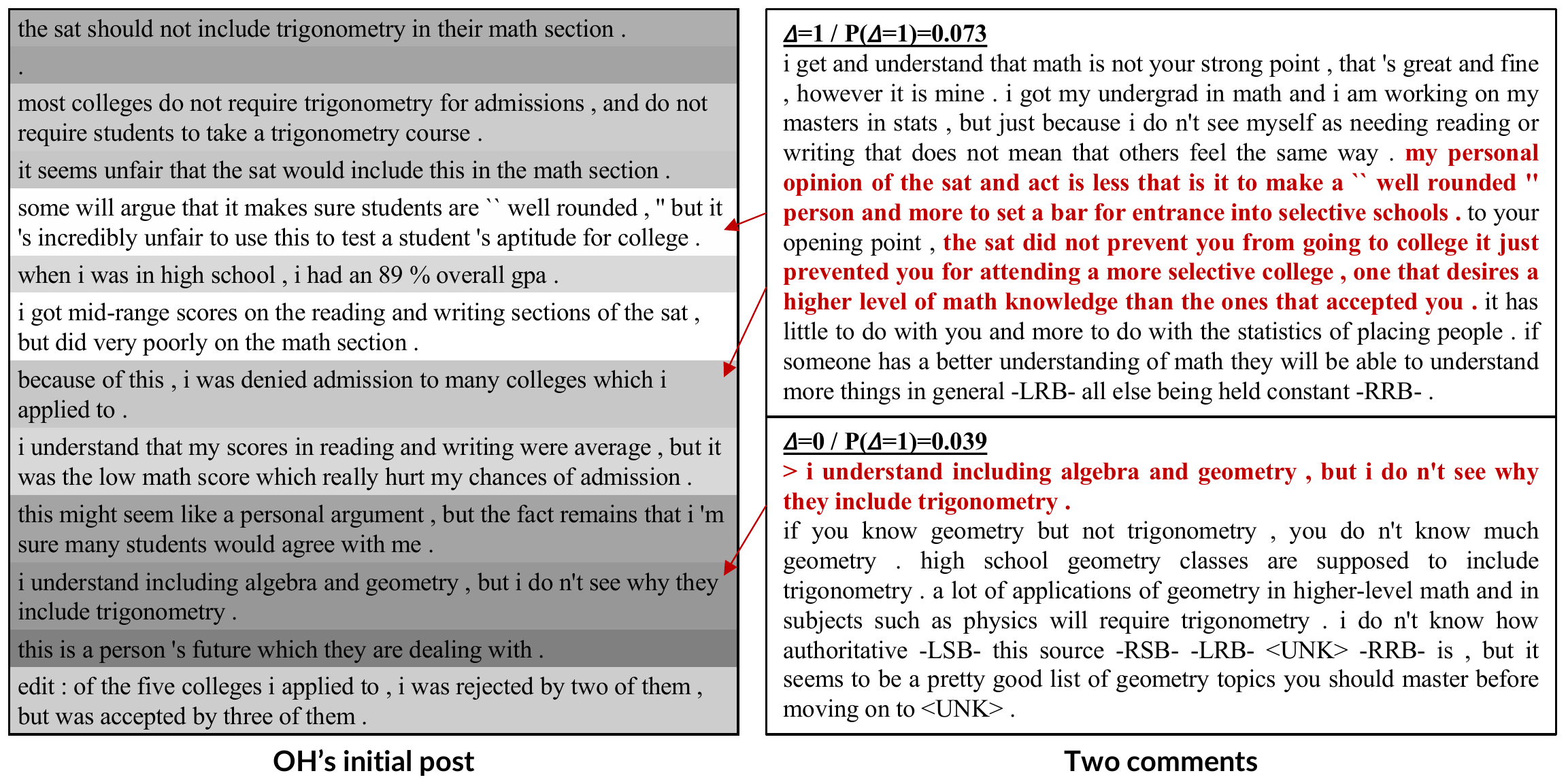}
    \caption{Example discussion with the OH's initial post (left), a successful comment (top right), and an unsuccessful comment (bottom right). The OH's post is colored based on attention weights (the higher attention the brighter). Sentences with college and SAT sections (\emph{reading}, \emph{writing}, \emph{math}) get more attention than sentences with other subjects (\emph{algebra}, \emph{geometry}). The successful comment addresses parts with high attention, whereas the unsuccessful comment addresses parts with low attention.\label{fig:attention_example}}
\end{figure*}

\paragraph{RQ2. Can the model identify vulnerable sentences, which are more likely to change the OH's view when addressed? If so, what properties constitute vulnerability?} 
Our rationale behind vulnerable region detection is that the model is able to learn to pay more attention to sentences that are more likely to change the OH's view when addressed. If the model successfully does this, then we expect more alignment between the attention mechanism and sentences that are actually addressed by successful comments that changed the OH's view.

To verify if our model works as designed, we randomly sampled 30 OH posts from the test set, and for each post, the first successful and unsuccessful comments. We asked a native English speaker to annotate each comment with the two most relevant sentences that it addresses in the OH post, without knowledge of how the model computes vulnerability and whether the comment is successful or not.

After this annotation, we computed the average attention weight of the two selected sentences for each comment. We ran a paired sample $t$-test and confirmed that the average attention weight of sentences addressed by successful comments was significantly greater than that of sentences addressed by unsuccessful comments ($p < 0.05$). Thus, as expected in the case where the attention works as designed, the model more often picks out the sentences that successful challengers address.

As to what the model learns as vulnerability, in most cases, the model attends to sentences that are not punctuation marks, bullet points, or irrelevant to the topic (e.g., \emph{can you cmv?}). A successful example is illustrated in Figure \ref{fig:attention_example}.  More successful and unsuccessful examples are included in Appendix \ref{app:vulnerability_examples}.

\paragraph{RQ3. What kinds of interactions between arguments are captured by the model?}
We first use existing argumentation theories as a lens for interpreting interaction embeddings (refer to Section \ref{sec:background}). For this, we sampled 100 OH posts with all their comments and examined the 150 sentence pairs that have the highest value for each dimension of the interaction embedding (the dimensionality of interaction embeddings is 3 for the best performing configuration). 22\% of the pairs in a dimension capture the comment asking the OH a question, which could be related to shifting the burden of proof. In addition, 23\% of the top pairs in one dimension capture the comment pointing out that the OH may have missed something (e.g., \emph{you don't know the struggles ...}). This might represent the challengers' attempt to provide premises that are missing in the OH's reasoning.

As providing missing information plays an important role in our data, we further examine if this attempt by challengers is captured in interaction embeddings even when it is not overtly signaled (e.g., \emph{You don't know ...}). We first approximate the novelty of a challenger's information with the topic similarity between the challenger's sentence and the OH's sentence, and then see if there is a correlation between topic similarity and each dimension of interaction embeddings (details are in Appendix \ref{app:topic_similarity}). As a result, we found only a small but significant correlation (Pearson's $r = -0.04$) between topic similarity with one of the three dimensions.

Admittedly, it is not trival to interpret interaction embeddings and find alignment between embedding dimensions and argumentation theories. The neural network apparently learns complex interactions that are difficult to interpret in a human sense. It is also worth noting that the top pairs contain many duplicate sentences, possibly because the interaction embeddings may capture sentence-specific information, or because some types of interaction are determined mainly by one side of a pair (e.g., disagreement is manifested mostly on the challenger's side).

\paragraph{TFIDF} 
We examine successful and unsuccessful styles reflected in TFIDF-weighted $n$-grams, based on their weights learned by logistic regression (top $n$-grams with the highest and lowest weights are in Appendix \ref{app:tfidf}). First, challengers are more likely to change the OH's view when talking about themselves than mentioning the OH in their arguments. For instance, first-person pronouns (e.g., \emph{i} and \emph{me}) get high weights, whereas second-person pronouns (e.g., \emph{you\_are} and \emph{then\_you}) get low weights. Second, different kinds of politeness seem to play roles. For example, markers of negative politeness (\emph{can} and \emph{can\_be}, as opposed to \emph{should} and \emph{no}) and negative face-threatening markers (\emph{thanks}), are associated with receiving a $\Delta$. Third, asking a question to the OH (e.g., \emph{why}, \emph{do\_you}, and \emph{are\_you}) is negatively associated with changing the OH's view.

\section{Conclusion\label{sec:conclusion}}
We presented the Attentive Interaction Model, which predicts an opinion holder (OH)'s change in view through argumentation by detecting vulnerable regions in the OH's reasoning and modeling the interaction between the reasoning and a challenger's argument. According to the evaluation on discussions from the Change My View forum, sentences identified by our model to be vulnerable were addressed more by successful challengers than by unsuccessful ones. The model also effectively captured interaction information so that both vulnerability and interaction information increased accuracy in predicting an OH's change in view.

One key limitation of our model is that making a prediction based only on one comment is not ideal because we miss context information that connects successive comments. As a discussion between a challenger and the OH proceeds, the topic may digress from the initial post. In this case, detecting vulnerable regions and encoding interactions for the initial post may become irrelevant. We leave the question of how to transfer contextual information from the overall discussion as future work.

Our work is a step toward understanding how to model argumentative interactions that are aimed to enrich an interlocutor's perspective. Understanding the process of productive argumentation would benefit both the field of computational argumentation and social applications, including cooperative work and collaborative learning.

\section*{Acknowledgments}
This research was funded by the Kwanjeong Educational Foundation and NSF grants IIS-1546393, ACI-1443068, and DGE-1745016. The authors thank Amazon for their contribution of computation time through the AWS Educate program.

\bibliography{naaclhlt2018}
\bibliographystyle{acl_natbib}

\appendix

\section{Implementation Details\label{app:implementation}}

\subsection{Topics in the Data}
Topics are extracted using \textsf{LatentDirichletAllocation} in \textsf{scikit-learn v0.19.1}, with the following setting:
\begin{itemize}
\setlength\itemsep{0em}
    \item \texttt{n\_components}: 20
    \item \texttt{max\_iter}: 200
    \item \texttt{learning\_method}: online
    \item \texttt{learning\_offset}: 50
\end{itemize}

\subsection{AIM}
We implemented our model in \textsf{PyTorch 0.3.0}.

\subsection{Baseline}
We use \textsf{LogisticRegression} in \textsf{scikit-learn v0.19.1}, with the default settings.

\subsection{TFIDF Features}
TFIDF is extracted using \textsf{TfidfVectorizer} in \textsf{scikit-learn v0.19.1}, with the default setting.

\section{Data Preprocessing\label{app:data}}
In the CMV forum, DeltaBot replies to an OH's comment with the confirmation of a $\Delta$, along with the user name to which the OH replied.
For most OH replies, the (non-)existence of a $\Delta$ indicates whether a comment to which the OH replied changed the OH's view. However, an OH's view is continually influenced as they participate in argumentation, and thus a $\Delta$ given to a comment may not necessarily be attributed to the comment itself. One example is when a comment does not receive a $\Delta$ when the OH reads it for the first time, but the OH comes back and gives it a $\Delta$ after they interact with other comments. In such cases, we may want to give a credit to the comment that actually led the OH to reconsider a previous comment and change the view.

Hence, we use the following labeling that considers the order in which OHs read comments. We treat the (non-)existence of a $\Delta$ in an OH comment as a label for the last comment that the OH read. We reconstruct the order in which the OH reads comments as follows. We assume that when the OH writes a comment, he/she has read all prior comments in the path to that comment. 

Based on this assumption, we linearize (i.e., flatten) the original tree structure of the initial post and all subsequent comments into a linear sequence $S$. Starting with empty $S$, for each of the OH's comments in chronological order, its ancestor comments that are yet to be in $S$ and the comment itself are appended to $S$. And for each of the OH's comments, its preceding comment in $S$ is labeled with $\Delta = 1$ if the OH's comment has a $\Delta$ and 0 otherwise.

This ensures that the label of a comment to which the OH replied is the (non-)existence of a $\Delta$ in the OH's first reply. If an OH reply is not the first reply to a certain comment (as in the scenario mentioned above), or a comment to which the OH replied is missing, the (non-)existence of a $\Delta$ in that reply is assigned to the comment that we assume the OH read last, which is located right before the OH's comment in the restructured sequence.

\section{Vulnerability Examples\label{app:vulnerability_examples}}

\begin{figure*}
    \centering
    \includegraphics[width=\linewidth]{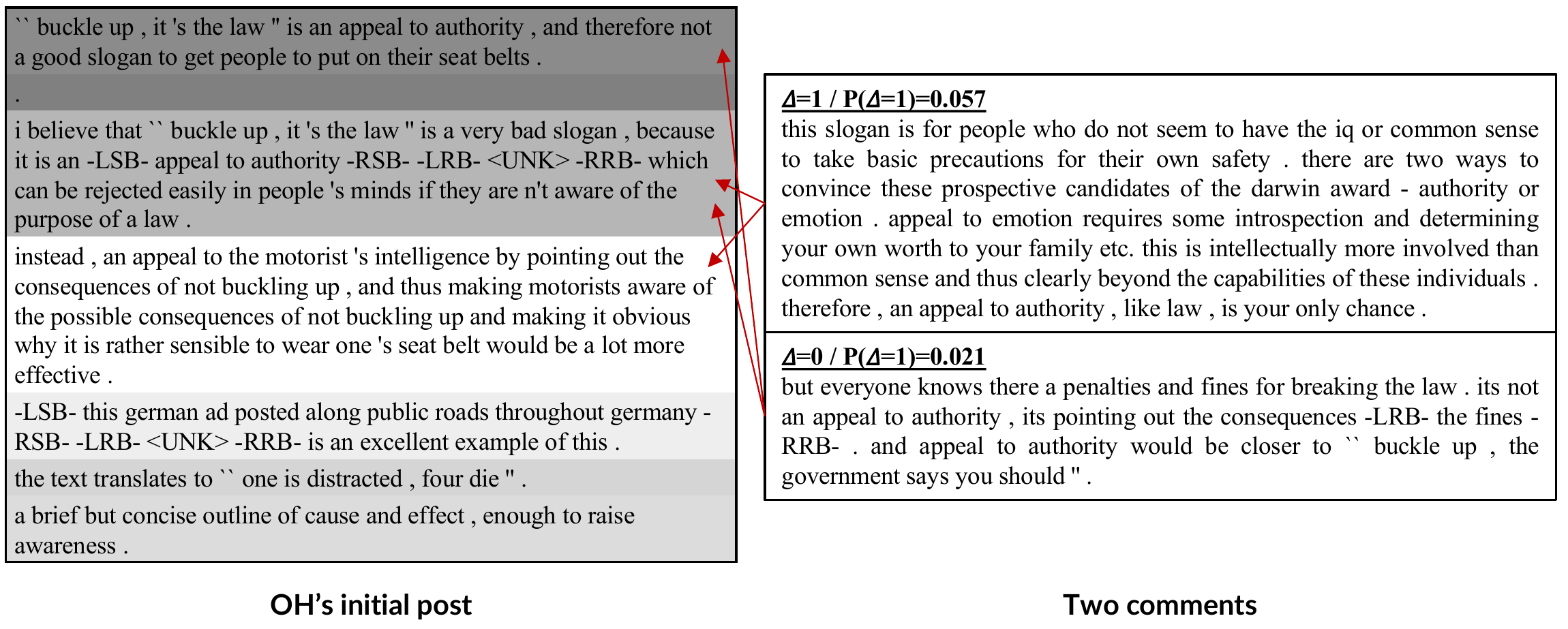}
    \includegraphics[width=\linewidth]{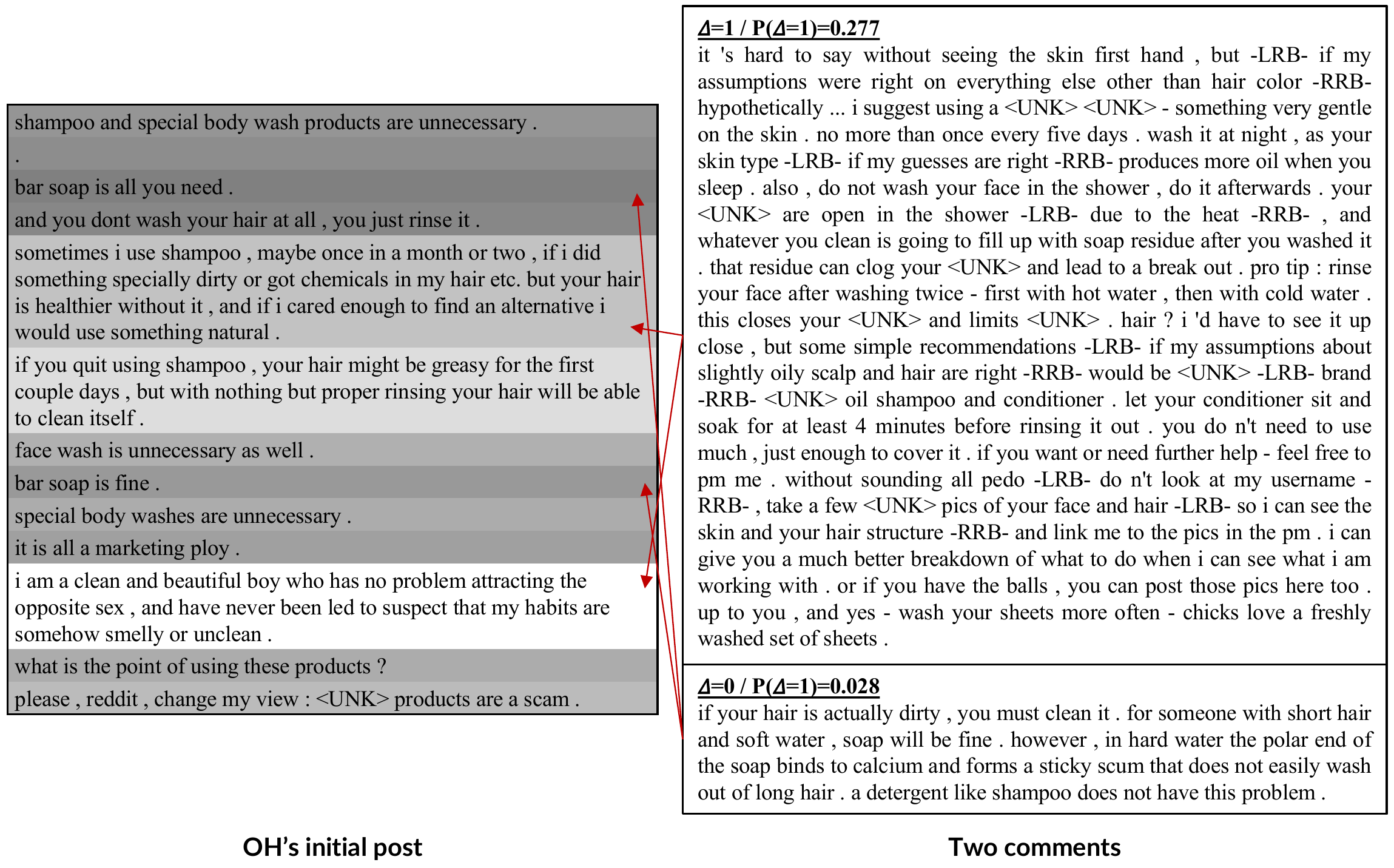}
    \caption{Successful examples of vulnerable region detection.\label{fig:good_examples}}
\end{figure*}

\begin{figure*}
    \centering
    \includegraphics[width=\linewidth]{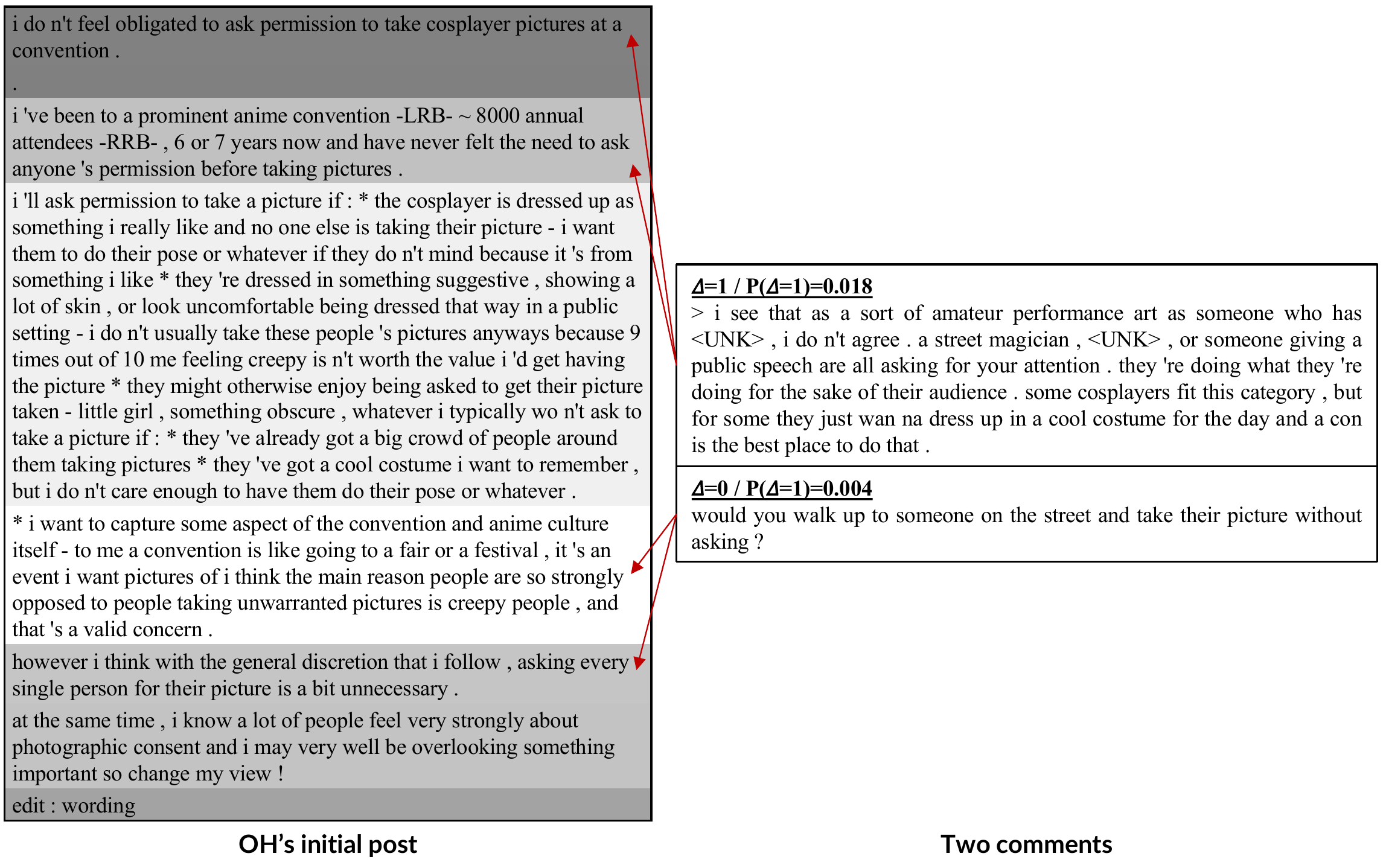}
    \includegraphics[width=\linewidth]{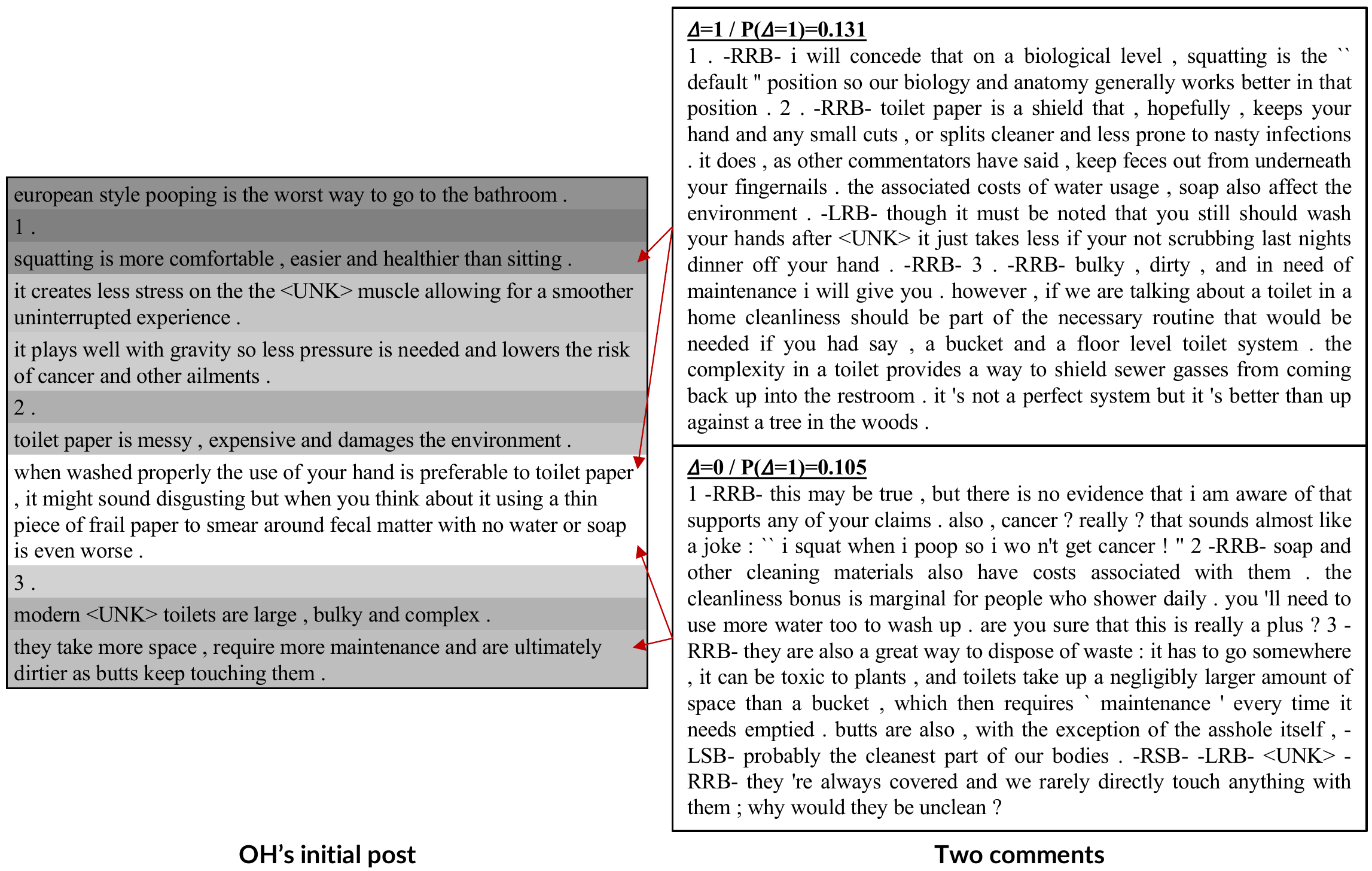}
    \caption{Unsuccessful examples of vulnerable region detection.\label{fig:bad_examples}}
\end{figure*}

Figure \ref{fig:good_examples} and Figure \ref{fig:bad_examples} show successful and unsuccessful examples of vulnerable region detection. All examples are from the test set. 

\section{Topic Similarity between Sentences\label{app:topic_similarity}}
The topic similarity between a pair of sentences is computed as the consine similarity between the topic distributions of the sentences. 

The first step is to extract topics. Using \textsf{LatentDirichletAllocation} in \textsf{scikit-learn v0.19.1}, we ran LDA on the entire data with 100 topics, taking each post/comment as a document. We treat the top 100 words for each topic as topic words.

The second step is to compute the topic distribution of each sentence. We simply counted the frequency of occurrences of topic words for each topic, and normalized the frequencies across topics.

Lastly, we computed the cosine similarity between the topic distributions of a pair of sentences.

\section{Top TFIDF $n$-grams\label{app:tfidf}}
The $n$-grams that contribute most to $\Delta$ prediction for logistic regression are shown in table \ref{tab:lr_features}.

\begin{table}[t]
	\centering
    \begin{tabularx}{\linewidth}{XX} \toprule
    $n$-grams for $\Delta = 1$ & $n$-grams for $\Delta = 0$ \\ \midrule
 	and, in, for, use, it, on, thanks, often, delta, time, depression, -RRB-, lot, -LRB-, or, i, can, \&, with, more, as, band, *, \#, me, - LRB-\_-RRB-, can\_be, has, deltas, when & ?, $>$,  sex, why, do\_you, wear, relationship, child, are\_you, op, mother, should, wearing, teacher, then, it\_is, same, no, circumcision, you\_are, then\_you, baby, story \\ \bottomrule
	\end{tabularx}
    \caption{Top $n$-grams with the most positive/negative weights for logistic regression.\label{tab:lr_features}}
\end{table}

\end{document}